\documentclass[sigconf,preprint,nonacm]{acmart}

\AtBeginDocument{%
  }

\setcopyright{none}

\microtypesetup{expansion=false}
\usepackage{hyperref}
\usepackage{subcaption}

\begin{document}
\title{Low-Cost Teleoperation Extension for Mobile Manipulators}


\author{Danil Belov}
\authornote{*These authors contributed equally to this work.}
\affiliation{
  \institution{Research Center for Digital Engineering and Innovation}
  \city{Moscow}
  \country{Russia}
}
\email{d.belov@rcdei.com}

\author{Artem Erkhov}
\authornotemark[1]
\affiliation{
  \institution{Research Center for Digital Engineering and Innovation}
  \city{Moscow}
  \country{Russia}
}
\email{a.erkhov@rcdei.com}

\author{Yaroslav Savotin}
\affiliation{
  \institution{Research Center for Digital Engineering and Innovation}
  \city{Moscow}
  \country{Russia}
}
\email{y.savotin@rcdei.com}

\author{Tatiana Podladchikova}
\affiliation{
  \institution{Research Center for Digital Engineering and Innovation}
  \city{Moscow}
  \country{Russia}
}
\email{t.podladchikova@rcdei.com}

\author{Pavel Osinenko}
\affiliation{
  \institution{Research Center for Digital Engineering and Innovation}
  \city{Moscow}
  \country{Russia}
}
\email{p.osinenko@rcdei.com}

\author{Dzmitry Tsetserukou}
\affiliation{
  \institution{Research Center for Digital Engineering and Innovation}
  \city{Moscow}
  \country{Russia}
}
\email{d.tsetserukou@rcdei.com}


\begin{abstract}
Teleoperation of mobile bimanual manipulators requires simultaneous control of high-dimensional systems, often necessitating expensive specialized equipment. We present an open-source teleoperation framework that enables intuitive whole-body control using readily available commodity hardware. Our system combines smartphone-based head tracking for camera control, leader arms for bilateral manipulation, and foot pedals for hands-free base navigation. Using a standard smartphone with IMU and display, we eliminate the need for costly VR helmets while maintaining immersive visual feedback. The modular architecture integrates seamlessly with the XLeRobot framework, but can be easily adapted to other types of mobile manipulators. We validate our approach through user studies that demonstrate improved task performance and reduced cognitive load compared to keyboard-based control.
\end{abstract}





\maketitle

\begin{figure}[t]
    \centering
    \begin{subfigure}[t]{0.48\columnwidth}
        \centering
        \includegraphics[width=\textwidth]{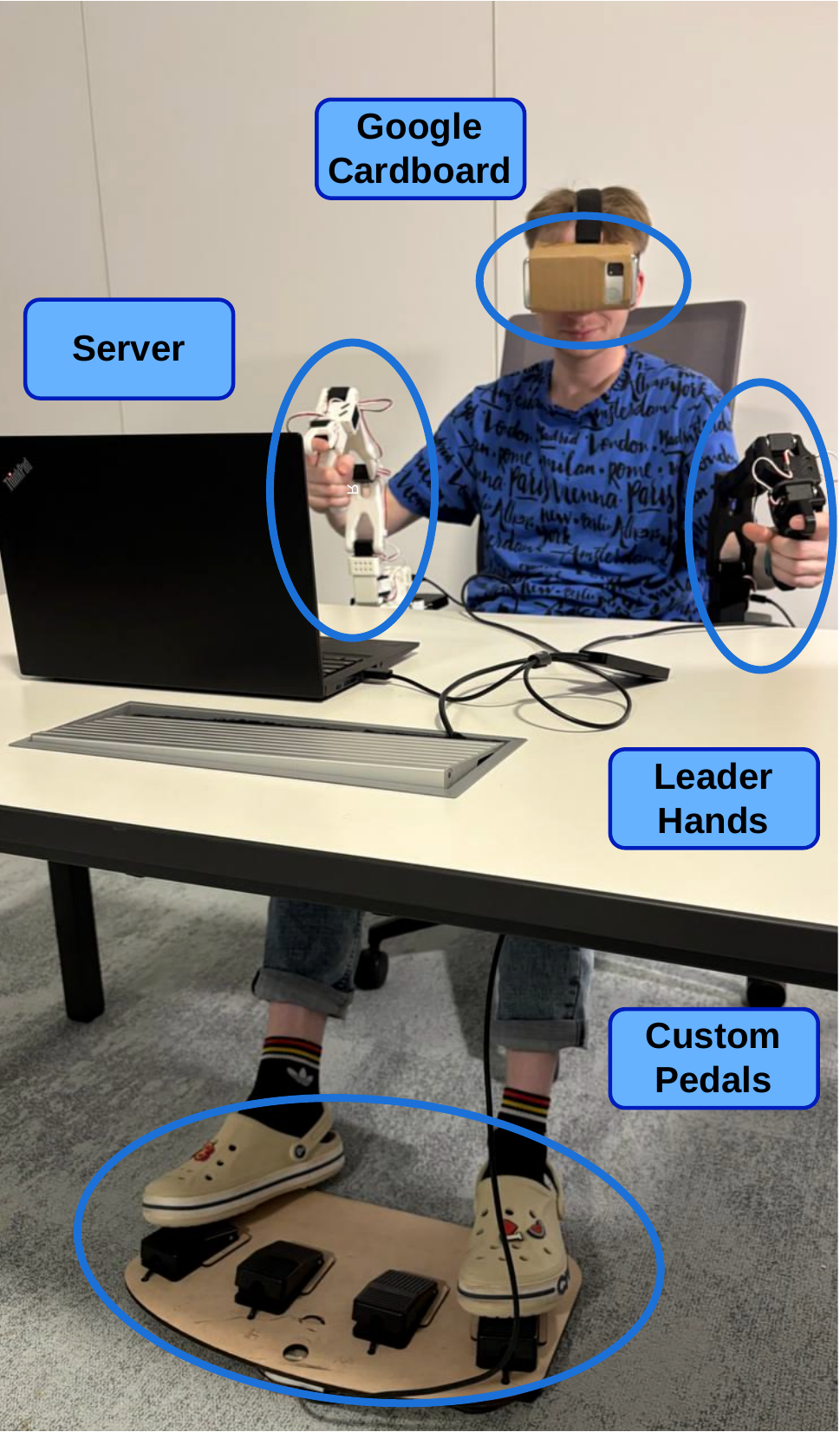}
        \caption{Teleoperation Setup.}
        \label{fig:vr_setup}
    \end{subfigure}
    \hfill
    \begin{subfigure}[t]{0.48\columnwidth}
        \centering
        \includegraphics[width=\textwidth]{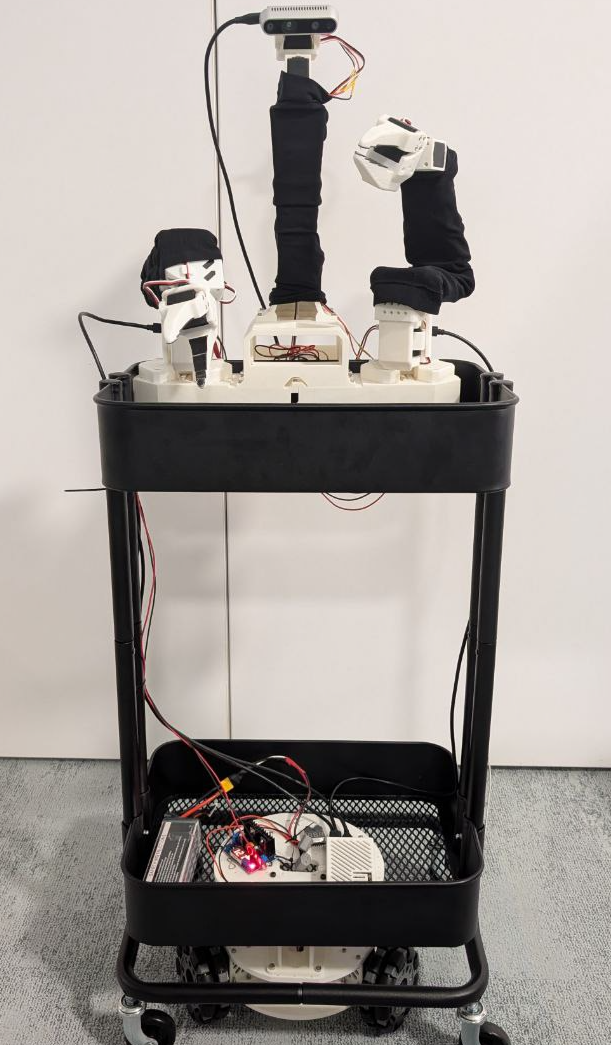}
        \caption{Mobile robot.}
        \label{fig:mobile_platform}
    \end{subfigure}
    \caption{Manipulation system components.}
    \label{fig:system_overview}
\end{figure}

\section{Introduction}

Mobile manipulators combine mobility with dexterity, enabling robots to navigate environments while performing manipulation tasks. However, effectively controlling these high-dimensional systems remains challenging, as operators must simultaneously coordinate base locomotion, arm movements, and camera orientation across numerous degrees of freedom.

Existing teleoperation solutions face fundamental trade-offs. High-end systems using VR headsets and motion capture provide immersive control but require expensive specialized hardware, creating barriers for research groups. More affordable keyboard-based interfaces offer discrete control that conflicts with natural manipulation, forcing operators to mentally translate desired motions into sequences of key presses while their hands are occupied with input devices.

We address these limitations by leveraging commodity hardware: smartphones for head tracking and visual feedback and foot pedals for hands-free base navigation. The smartphone serves dual purposes—its IMU tracks head orientation while its display provides immersive visual feedback—eliminating the need for costly VR equipment. Foot pedals free the hands for manipulation, creating clear separation between navigation and manipulation control.

Built on the LeRobot framework \cite{cadene2024lerobot}, our system provides seamless integration for real-world data collection and policy training. Through user studies, we demonstrate improved task performance and reduced cognitive load compared to keyboard-based control.

Our contributions include:
\begin{enumerate}
\item A complete open-source teleoperation framework combining smartphone-based head control with adaptive visual feedback, two arm manipulation, and pedal-based navigation
\item Empirical validation that demonstrate improved performance and reduced cognitive load compared to baseline approaches
\item Integration with LeRobot supporting real-world deployment and simulation-based development
\end{enumerate}

The complete system implementation is available at \url{https://github.com/HRI2026LBR/xlerobot-teleop}.

\section{Related Works}

Teleoperation of mobile manipulators has been extensively studied, with various approaches addressing the challenges of simultaneous control of multiple degrees of freedom. Research in human factors and ergonomics has demonstrated that simultaneous control of multiple degrees of freedom creates significant cognitive burden \cite{ergonomics2013}, and interfaces requiring constant modality switching lead to decreased performance and mental fatigue. This aligns with findings from Wizard of Oz interfaces in HRI research \cite{bejarano2024woz}, which emphasize the importance of seamless, intuitive control for effective human-robot interaction. This section reviews the evolution of teleoperation interfaces and highlights the remaining challenges that motivate our approach.

Early approaches to whole-body teleoperation emphasized physical coupling between operator and robot. Mobile ALOHA \cite{mobilealoha2024} enables natural coordination by physically tethering the operator to directly backdrive the base wheels while puppeteering the manipulators. This approach achieves intuitive motion mapping but sacrifices remote operation capabilities and causes operator fatigue during extended use.

To enable remote control, researchers turned to head-based interfaces. Padmanabha et al. \cite{hat2023} introduced a head-worn IMU embedded in a baseball cap, mapping head orientation to velocity commands for base, arm, and end-effector. While providing accessibility and remote operation, this approach requires frequent calibration and suffers from drift due to limited sensor fusion capabilities. Building on head-based control, VR systems \cite{opentelevision2024}, \cite{viewvr}, \cite{galarza2023} enhance spatial perception through stereoscopic video streaming and active camera control that follows operator head movements. These immersive interfaces improve manipulation precision but introduce significant cost barriers through specialized VR hardware requirements.

Recognizing the limitations of single-modality interfaces, modular frameworks emerged to unify multiple input devices. TeleMoMa \cite{telemoma2024} integrates RGB-D cameras, VR controllers, keyboards, and joysticks within a flexible architecture, while Sketch-MoMa \cite{sketchmoma2025} explores vision-language interpretation of hand-drawn sketches for task generation. These multimodal approaches lower entry barriers but often compromise precision or introduce processing latency that conflicts with real-time manipulation requirements.

As an alternative to requiring simultaneous control of all degrees of freedom, automation-based approaches delegate navigation to reduce operator burden. Honerkamp et al. \cite{momateleop2024} present MoMa-Teleop, where base motion is handled by a pre-trained reinforcement learning agent while the operator focuses solely on end-effector control. This significantly reduces cognitive load but requires extensive pre-training and limits flexibility for non-standard scenarios.

Despite this evolution from physical tethering to head interfaces, VR systems, modular frameworks, and automation-assisted control, a fundamental challenge remains: enabling simultaneous, intuitive control of arms, head, and base using affordable hardware without compromising precision or requiring extensive training. Existing systems face inherent trade-offs between cost and capability, between physical presence and remote operation, and between automation and flexibility. The challenge of coordinating multiple control inputs—manipulator arms, active camera system, and mobile base—continues to create high cognitive burden, leading to suboptimal performance or extensive training requirements.

Our system addresses these limitations by combining commodity hardware in a modular architecture that separates control modalities without requiring automation. Leader-follower arms provide intuitive bilateral manipulation, smartphone-based head control offers VR-like immersion without expensive hardware, and pedal-based navigation enables hands-free base control. This design philosophy maintains direct operator control across all degrees of freedom while reducing cognitive load through ergonomic modality separation. This approach emerges as a natural response to the accumulated challenges identified in existing teleoperation approaches.

\section{System Overview}

The XLeRobot \cite{wang2025xlerobot} is a mobile bimanual robot platform designed for general manipulation tasks. The system consists of a three-wheeled omnidirectional mobile base, two 5-DOF manipulator arms (left and right), and a 2-DOF head mechanism. The robot is equipped with a single head-mounted camera that provides first-person perspective for the operator.

The baseline teleoperation system employs a bimanual leader-follower configuration using two SO101 leader arms. The leader arms are physically controlled by the operator and capture joint positions in real-time, which are then transmitted to the corresponding follower arms on the XLeRobot. This bilateral teleoperation approach provides intuitive and precise control of the robot's manipulators, allowing the operator to perform complex manipulation tasks through direct physical interaction with the leader devices.

The system also includes built-in keyboard-based control for the head and mobile base. Keyboard commands allow control of head orientation (yaw and roll) and base movement (forward/backward, lateral translation, and rotation) through key presses. However, this approach has significant limitations: (1) keyboard input provides discrete, step-based control rather than continuous motion, resulting in jerky and unnatural movements; (2) the operator's hands are occupied with keyboard input, conflicting with the simultaneous use of leader arms for manipulation; (3) the control lacks the intuitive mapping between operator motion and robot motion that is essential for immersive teleoperation. These limitations motivated the development of the enhanced control modalities described below.

\subsection{Smartphone-Based Head Control}

We extend the existing teleoperation framework with two key enhancements: (1) smartphone-based head control and adaptive visual feedback, and (2) pedal-based base control. These additions enable more natural and immersive teleoperation while maintaining compatibility with the existing arm control system.

The head control subsystem utilizes a smartphone as a wireless IMU sensor and VR display. The smartphone's built-in orientation sensors (gyroscope, accelerometer, and magnetometer) provide real-time head orientation data through a WebSocket connection over HTTPS. The orientation angles (roll, pitch, and yaw) are normalized to a ±90° range relative to an initial calibration point, ensuring stable control within the robot's mechanical limits. The system implements automatic calibration upon connection, establishing the initial head position as the zero reference point, with the capability for dynamic recalibration during operation.

In our user study, we tested the Smartphone VR setup using an iPhone 16 (170 grams) with Google Cardboard VR \cite{googlecardboard2024} (80 grams), resulting in a total weight of 250 grams, compared to Meta Quest 3 (515 grams). This significant weight difference contributes to reduced neck fatigue during extended teleoperation sessions.

\begin{figure}[h]
    \centering
    \includegraphics[width=\linewidth]{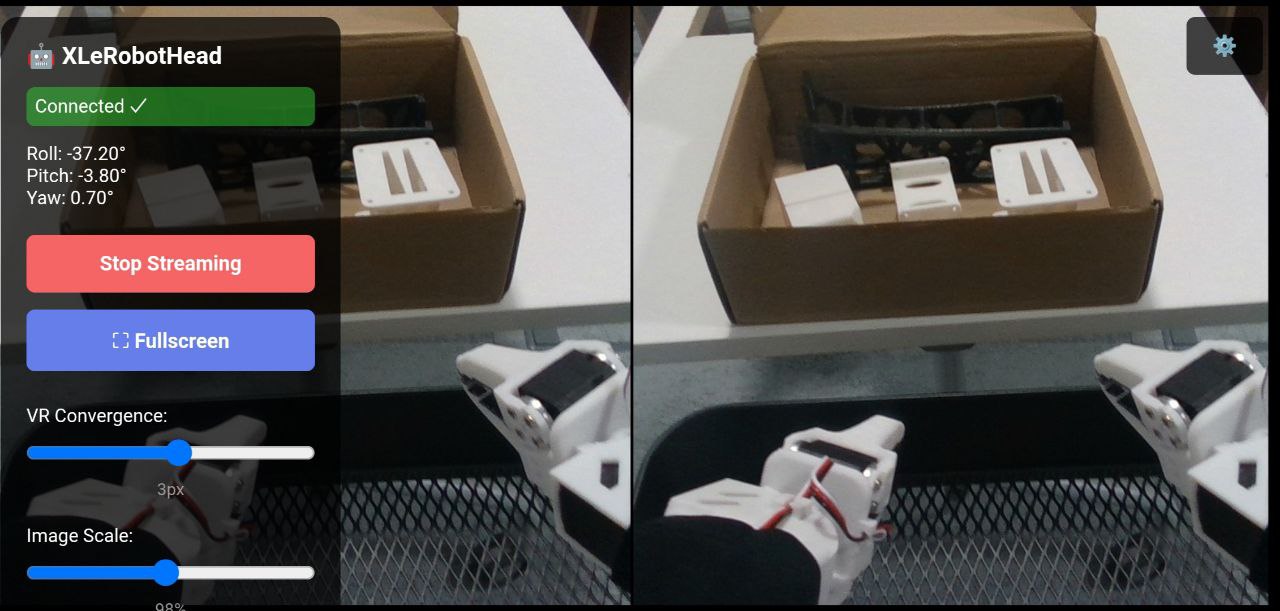}
    \caption{Smartphone VR interface control.}
    \label{fig:interface}
\end{figure}

The visual feedback system streams the robot's head camera feed to the smartphone in real-time, presenting it in a split-screen VR format optimized for head-mounted displays (\autoref{fig:interface}). The system includes several adaptive features to enhance the operator experience:

\begin{itemize}
    \item Full-screen display: The video stream is rendered without borders, utilizing the entire smartphone screen for maximum immersion.
    \item Convergence adjustment: A configurable slider allows fine-tuning of the inter-eye image offset to eliminate double vision and match the operator's interpupillary distance.
    \item Image scaling: A zoom control enables adjustment of the displayed image size to accommodate different VR headset form factors and user preferences.
\end{itemize}

These parameters are persisted in browser local storage, maintaining user preferences across sessions. The video stream is optimized for VR viewing, automatically rotating portrait-oriented frames to landscape and maintaining aspect ratio. Maximum resolution is up to 1920$\times$1080 pixels.

\subsection{Pedal-Based Base Control}

A four-pedal interface provides hands-free base navigation. Pedals correspond to forward, backward, left, and right translation. Rotation is achieved by combining the forward pedal with the left (counterclockwise) or right (clockwise) pedal.

An STM32 microcontroller processes pedal signals, performing debouncing and state management before sending velocity commands to the base. This design frees the operator's hands for manipulation, offers an ergonomic control method, and reduces cognitive load by separating navigation from arm control.

\subsection{System Integration}

The system integrates leader arms, smartphone head control, and pedal base control into a unified 30 Hz control loop. Each modality operates independently but combines into a single command. The architecture is modular, supports disconnections, and uses continuous calibration to prevent drift during extended operation.

\section{User Study}

\textbf{Participants.} A total of 30 participants (aged 24–35) were recruited for the study. No dedicated pre-training was provided prior to the experiment. Only two participants reported prior experience with robotic systems, while the remaining participants had little or no exposure to robotics.

\textbf{Tasks}. The user study consisted of three representative manipulation tasks. These tasks involved a mobile robotic platform equipped with two manipulators operating in a laboratory environment:

1. Pack the 3d printed part in a box.

2. Place the bottle on the shelf in the locker (Fig.~\ref{fig:locker_task}).

\begin{figure}[h]
    \centering
    \begin{tabular}{@{}ccc@{}}
        \includegraphics[width=0.5\textwidth]{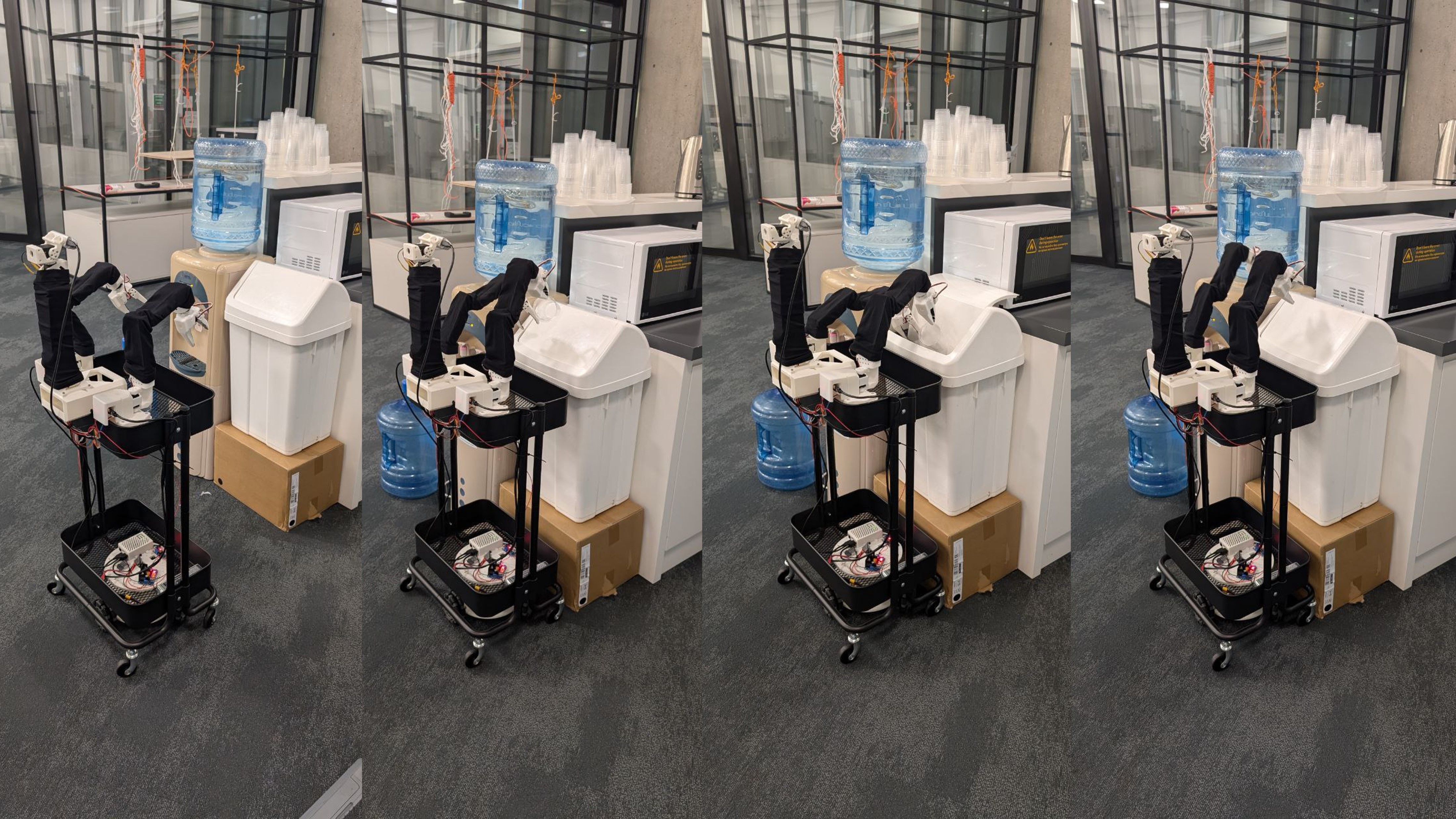} &
    \end{tabular}
    \caption{Trash can task sequence of actions.}
    \label{fig:trash_can_task}
\end{figure}

3. Throw the cup in the trash can (Fig.~\ref{fig:trash_can_task}).

Each participant performed the full sequence of three tasks using three different control setups: (1) conventional keyboard-based control interface, (2) Meta Quest 3 VR headset with pedals for base control, and (3) smartphone-based head control with pedals (the proposed system). The order of control setups was counterbalanced across participants to control for learning effects.

\begin{figure}[h]
  \centering
  \begin{tabular}{@{}ccc@{}}
      \includegraphics[width=0.5\textwidth]{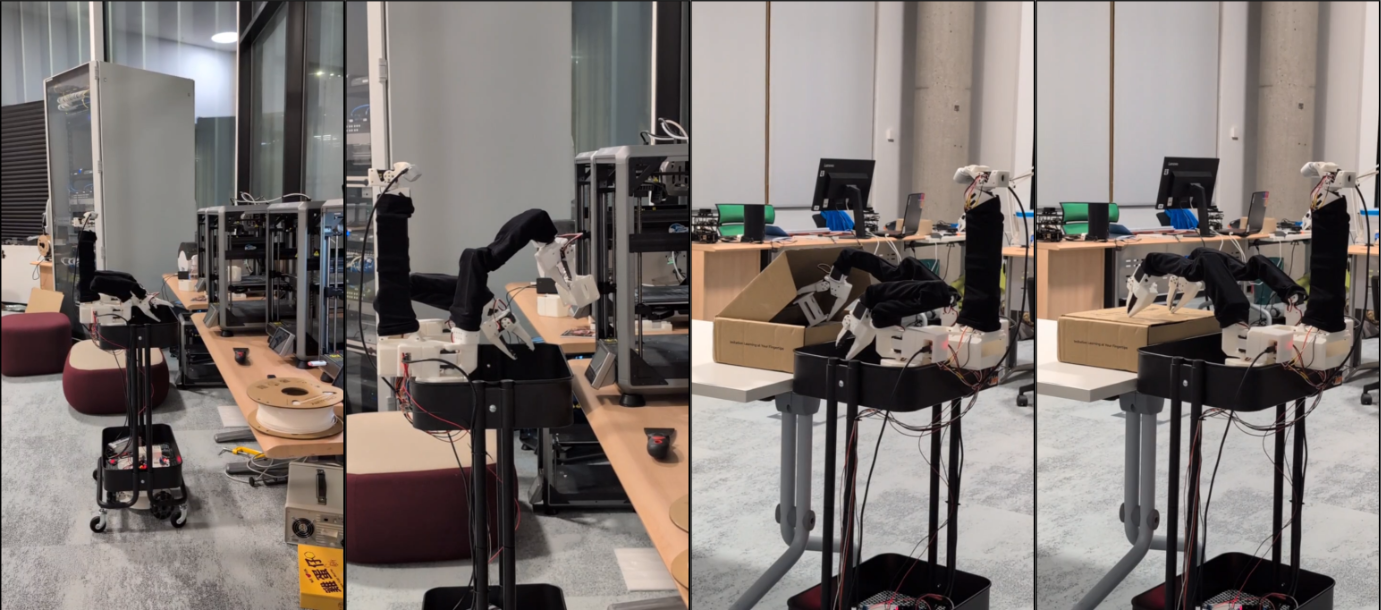} &
  \end{tabular}
  \caption{Pack printed part task sequence of actions.}
  \label{fig:locker_task}
\end{figure}

\begin{figure}[h]
    \centering
    \begin{tabular}{@{}ccc@{}}
        \includegraphics[width=0.5\textwidth]{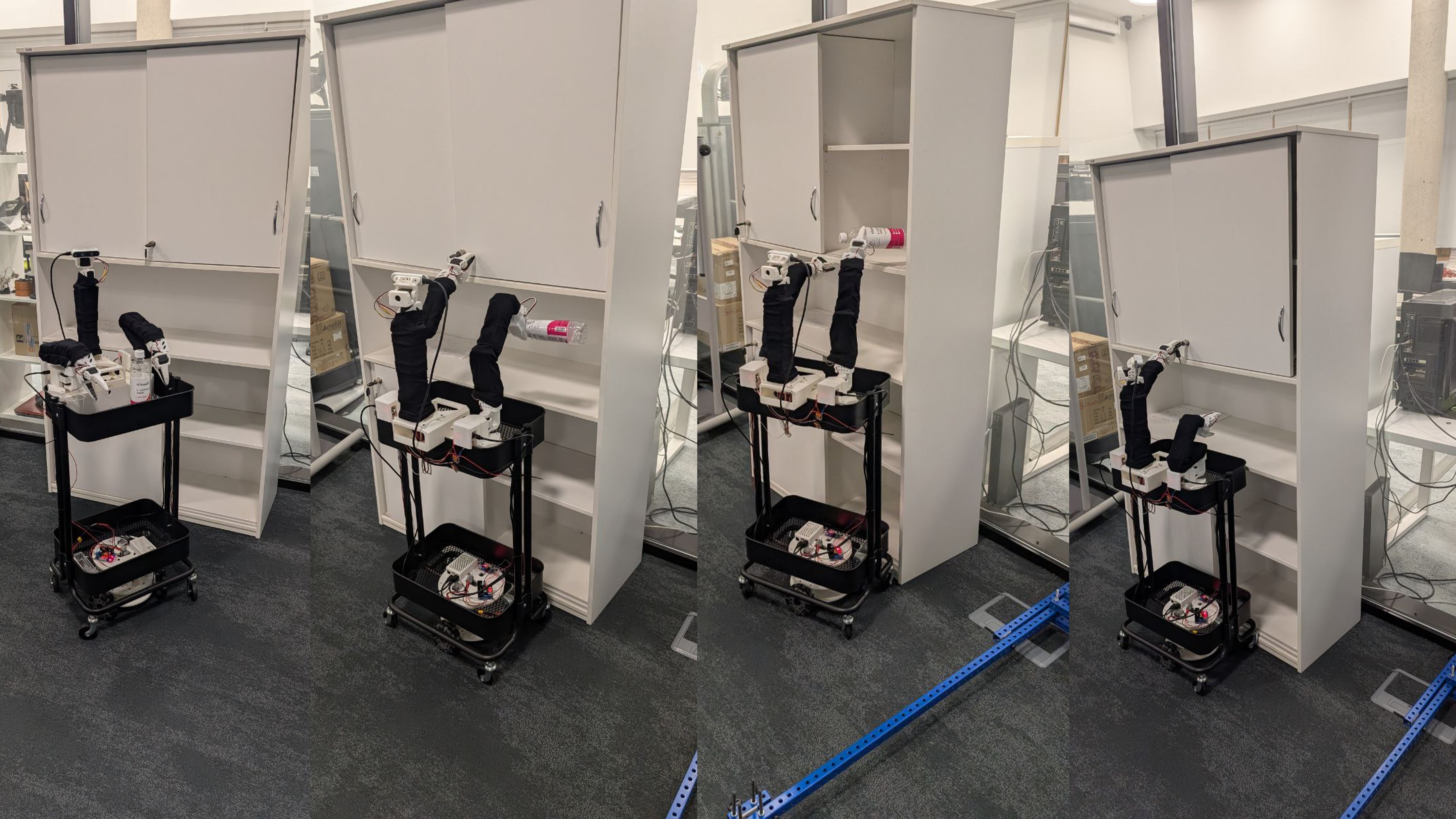} &
    \end{tabular}
    \caption{Locker task sequence of actions.}
    \label{fig:locker_task}
\end{figure}

\textbf{Measures.} Both objective performance metrics and subjective workload assessments were collected in order to evaluate the effectiveness and usability of the proposed low-cost teleoperation setup.

\textit{Objective measures} included task success rate and task completion time for each of the three tasks. A task was considered successful if the participant completed the required manipulation without external assistance.

\textit{Subjective measures} were collected using the NASA Task Load Index (NASA-TLX) questionnaire, administered after each control condition. The NASA-TLX evaluates perceived workload across six dimensions: Mental Demand, Physical Demand, Temporal Demand, Performance, Effort, and Frustration Level. Weighted overall workload scores were computed following the standard NASA-TLX procedure for all 30 participants 

\begin{table*}[h]
  \centering
  \caption{Overall task success rate and completion time across all participants and experiments.}
  \label{tab:overall_success}
  \begin{tabular}{lcccccc}
    \toprule
    \textbf{Control setup} &
    \multicolumn{2}{c}{\textbf{Pack part into a box}} &
    \multicolumn{2}{c}{\textbf{Place a bottle into a locker}} &
    \multicolumn{2}{c}{\textbf{Throw garbage into a garbage bin}} \\
    \cmidrule(lr){2-3} \cmidrule(lr){4-5} \cmidrule(lr){6-7}
    & \textbf{Success rate} & \textbf{Time} & \textbf{Success rate} & \textbf{Time} & \textbf{Success rate} & \textbf{Time} \\
    \midrule
    \textit{Keyboard setup}          & 65.5\% & 124 s & 43.6\% & 232 s & 75.3\% & 80 s \\
    \textit{Meta Quest 3 + pedals}  & 74.8\% & 108 s & 47.2\% & 210 s & 88.1\% & 64 s \\
    \textit{Smartphone VR + pedals}  & 74.4\% & 110 s & 46.6\% & 211 s & 89.6\% & 67 s \\
    \bottomrule
  \end{tabular}
  \vspace{1mm}
\end{table*}

\begin{figure*}[h]
    \centering
    \includegraphics[width=\textwidth]{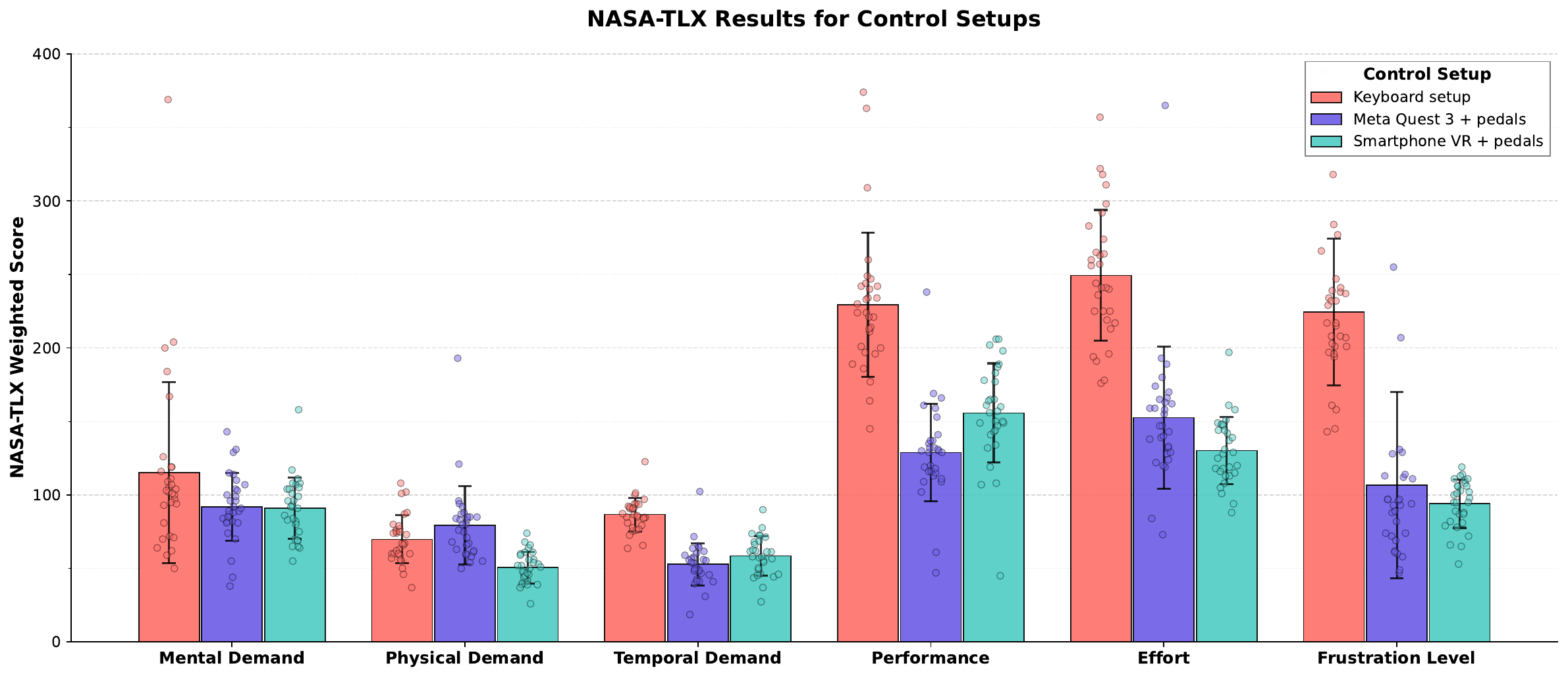}
    \caption{Taskload levels for Keyboard setup, Meta Quest 3 + pedals, and Smartphone VR + pedals.}
    \label{fig:NASA_TLX}
\end{figure*}

\textbf{Participants' Results.}
Objective performance results indicate that both VR-based teleoperation setups (Meta Quest 3 + pedals and Smartphone VR + pedals) consistently outperformed keyboard control across all tasks. As summarized in \autoref{tab:overall_success}, both VR setups resulted in higher task success rates and reduced average completion times compared to keyboard control. Meta Quest 3 + pedals achieved the best overall performance, with the most notable improvement observed in the trash can task, where it reached a success rate of 88.1\% (compared to 75.3\% for keyboard) and reduced average completion time from 80~s to 64~s. Smartphone VR + pedals also showed significant improvements, achieving 86.6\% success rate and 67~s completion time for the same task. These results suggest that continuous, simultaneous control of robot motion and manipulation contributes to more efficient task execution, with dedicated VR hardware providing additional performance benefits.

Subjective workload analysis using NASA-TLX further supports these findings in \autoref{fig:NASA_TLX}. The aggregated task load scores show a clear reduction in perceived workload when using both VR-based teleoperation systems compared to keyboard control. Across all participants, keyboard control resulted in consistently higher overall task load values, with many participants reporting high Effort, Frustration, and Temporal Demand. In contrast, both Meta Quest 3 + pedals and Smartphone VR + pedals showed lower and less variable scores, indicating improved usability and reduced cognitive burden. Qualitative feedback aligns with the quantitative workload measures. Participants reported that both VR-based teleoperation interfaces felt more natural and continuous, allowing them to observe the environment, navigate the robot, and manipulate objects simultaneously. In comparison, keyboard control required frequent mode switching between navigation and manipulation, which participants associated with increased mental demand and frustration. 

One participant with no prior robotics or VR experience (but with a driver's license) provided detailed feedback comparing the systems:

\begin{quote}
\itshape
It's much easier to control the robot's arms and move objects while looking into a VR device than at a computer monitor. Similarly, it's easier to control the robot's movement using pedals than via keyboard. It was intuitive for me.

Comparing Smartphone VR and Meta Quest 3, I noticed that my head got tired from the weight of Meta Quest 3. It's too heavy and my neck hurts. Smartphone VR is significantly lighter, although the picture quality is worse.
\end{quote}


\section{Conclusion}

The open-source implementation, built on the LeRobot framework and XLerobot mobile platform, lowers barriers to entry for mobile manipulation research. By proving that effective teleoperation can be achieved with readily available components, we hope to democratize access to advanced robotics capabilities. This enables broader participation in human-robot interaction research.

Future work will explore enhanced visual feedback through stereoscopic camera systems, continuous pedal interfaces for finer locomotion control, and integration of haptic feedback from the robot to the operator. Additionally, we plan to investigate how data collection using our system can accelerate imitation learning for autonomous mobile manipulation policies.

The complete system is available as open-source to support replication and extension by the research community.


\bibliographystyle{ACM-Reference-Format}
\bibliography{sample-base}

\end{document}